\documentclass[11pt]{article}

\usepackage[]{acl}

\usepackage[T1]{fontenc}
\usepackage[utf8]{inputenc}
\usepackage{times}
\usepackage{latexsym}
\usepackage{microtype}
\usepackage{inconsolata}

\usepackage{mathtools}
\usepackage{amssymb}
\usepackage{amsthm}
\usepackage{nicefrac}

\usepackage{graphicx}
\usepackage[table,xcdraw]{xcolor}
\usepackage{tikz}
\usepackage{epstopdf}
\usepackage{wrapfig}
\usepackage{float}

\usepackage{booktabs}
\usepackage{tabularx}
\usepackage{longtable}
\usepackage{makecell}
\usepackage{multirow}
\usepackage{colortbl}

\usepackage{caption}
\usepackage{subcaption}
\usepackage{placeins}

\usepackage{enumitem}
\usepackage{tcolorbox}
\tcbuselibrary{most}
\definecolor{darkred}{RGB}{139,0,0}
\newtcolorbox{AIbox}[1]{
  enhanced,
  colback=gray!3,
  colframe=black!55,
  boxrule=0.5pt,
  arc=1mm,
  left=1.5mm,
  right=1.5mm,
  top=1mm,
  bottom=1mm,
  fonttitle=\bfseries,
  fontupper=\small,
  title={#1}
}
\newcommand{\promptvar}[1]{\{\textcolor{darkred}{#1}\}}

\usepackage{pifont}
\usepackage{wasysym}

\usepackage{xspace}
\usepackage{fancyvrb}
\usepackage{listings}
\usepackage{lipsum}
\usepackage{titletoc}
\usepackage{lineno}

\usepackage{url}
\usepackage{hyperref}

\usepackage{lineno}
\usepackage[ruled,vlined]{algorithm2e}

\newcommand{\method}{\textsc{AdaMM}\xspace}

\usepackage{times}
\usepackage{latexsym}
\usepackage[normalem]{ulem}

\usepackage[T1]{fontenc}

\usepackage[utf8]{inputenc}

\usepackage{microtype}

\usepackage{inconsolata}

\usepackage{graphicx}

\title{Beyond Retrieval: Analytic Memory for Multimodal Agents}

\author{
 \textbf{Zhoujin Tian\textsuperscript{1}},
 \textbf{Hao Zhang\textsuperscript{2}},
 \textbf{Yao Tian\textsuperscript{2}}\thanks{Corresponding authors.},
 \textbf{Cheng Chen\textsuperscript{2}}
\\
 \textbf{Yakun Li\textsuperscript{2}},
 \textbf{Lei Zhang\textsuperscript{2}},
 \textbf{Xiaofang Zhou\textsuperscript{1}\footnotemark[1]}
\\
 \textsuperscript{1}HKUST,
 \textsuperscript{2}ByteDance
\\
   \small{\{ztianaf, zxf\}@cse.ust.hk} \\
   \small{\{zhanghao.ai, yao.tian, chencheng.sg, liyakun.hit, zhanglei.michael\}@bytedance.com}
}

\begin{document}
\maketitle

\begin{abstract}
Long-term multimodal memory must support not only retrieving relevant information but also computing over observations accumulated across interactions. Existing systems largely emphasize \emph{retrieval memory}, organizing interaction histories through summaries and indexes to return query-relevant information at multiple granularities, from high-level abstractions to underlying records. In this paper, we formulate \emph{analytic memory} as a complementary abstraction that organizes recurring multimodal observations into queryable structures supporting filtering, aggregation, ranking, and temporal comparison. We present \method, a framework that jointly supports retrieval and analytic memory. Rather than relying on application-defined schemas, \method extracts provenance-linked attribute-value observations from dialogue, images, and contextual metadata, discovers recurring field structures, and materializes them for analytical access. At inference time, a memory-aware planner decomposes queries into retrieval and analytic operations and routes each operation to the appropriate tools. Experiments on two long-term multimodal memory benchmarks, MemEye and MemGallery, show that \method improves performance by up to 11.3\% and 6.9\%, respectively.
\end{abstract}

\section{Introduction}
As Large Language Model (LLM) agents interact with users over increasingly long horizons, they accumulate extensive multimodal histories -- including dialogues, screenshots, images, documents, and contextual observations -- that quickly exceed the finite context windows of their backbone models \citep{webvoyager,visualwebarena,osworld,mementos}. Multimodal memory systems, which retain and organize such histories to support future reasoning~\citep{mirix,m2a}, have therefore become essential and  are widely adopted in applications such as personal assistance ~\citep{llmagents,m2a}, embodied interaction~\citep{palme,voyager}, and tool-augmented workflows~\citep{react,toolformer}.


\begin{figure}[t]
    \centering
\includegraphics[width=\linewidth]{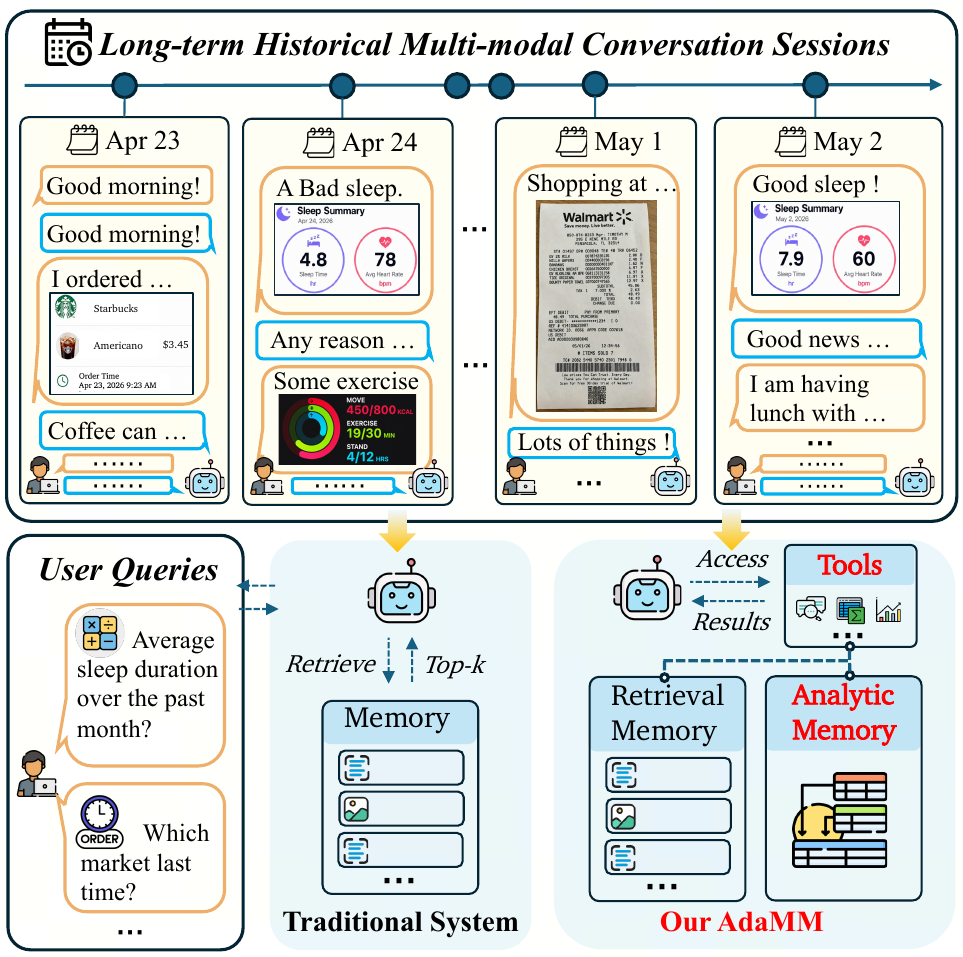}
    \caption{
    Long-term multimodal interaction histories contain observations distributed across sessions, modalities, and time. Answering user queries may require either retrieving specific past events or performing analytical operations, such as temporal selection and aggregation, over multiple observations.
    }
    \label{fig:intro}
\end{figure}


Recent multimodal memory systems have explored a broad range of designs, including textual abstraction~\citep{m2a}, specialized memory types~\citep{visualmem}, hybrid stores~\citep{mirix}, and cross-modal retrieval~\citep{mma}. 
Most existing systems follow a retrieve-then-answer paradigm, conditioning an LLM on a bounded set of relevant memories, which we denote as retrieval memory. Such systems are effective at selecting relevant memories from long interaction histories. However, long interaction histories also accumulate recurring observations that collectively form an append-only log. Analytical questions over such histories require complete, correctly scoped records and operations such as filtering, aggregation, ranking, and temporal selection. The relevance-based approach adopted by retrieval memory systems therefore creates a coverage--context trade-off: a small retrieval set may omit required observations, whereas a larger retrieval set consumes the limited context budget and introduces redundant or distracting evidence. For example, as illustrated in Figure~\ref{fig:intro}, computing a user's average sleep duration over the past month requires collecting all relevant measurements across multiple screenshots before aggregation; incomplete retrieval may therefore produce a biased estimate. 
We call this gap between relevance-based retrieval and analytic the \emph{retrieval--analysis mismatch}.

To bridge this gap, our key insight is that a memory system should couple two complementary subsystems: \emph{retrieval memory}, which identifies and supplies relevant historical records as context, and \emph{analytic memory}, which proactively organizes interaction histories into reusable and queryable structures and adaptively executes operations over them. Together, they support both flexible semantic recall and structured analysis, rather than requiring the LLM to repeatedly reconstruct the needed organization from retrieved records. Realizing analytic memory, however, presents two key challenges. First, interaction histories do not come with an explicit organizing schema, making it unclear what information should be preserved and how it should be structured to support future queries. Second, user queries rarely specify memory access strategies; the system therefore must reconcile what each query requires with what the evolving memory can
currently support. Even questions over the same underlying records may require fundamentally different operations (e.g., aggregation, ordering, filtering), posing challenges to the design of adaptive execution plans.


To address these challenges, we introduce \method, an  \textbf{Ada}ptive \textbf{M}ulti-view \textbf{M}emory framework that couples data-driven structure induction with memory-aware query planning. 
It maintains two complementary subsystems: \emph{retrieval memory} for flexible semantic access and \emph{analytic memory} for executable analysis over recurring observations. 
\method extracts provenance-linked attribute--value pairs from multimodal interactions (e.g., \textit{\{Sleep Time: 5.5 hrs\}}), induces recurrent schemas, and materializes them as structured tables, which are accessible via tools, while preserving unstructured evidence in a hierarchical semantic graph. 
At query time, a planner adaptively composes different analytic and retrieval tools based on the query and the memory structures currently available.  
This design enables grounded semantic recall and structured analysis over long-term interaction histories.

Our contributions are summarized as follows:
\begin{itemize}[leftmargin=*]
    \item We identify a \emph{retrieval--analysis mismatch} in long-term multimodal memory: retrieval-oriented interfaces support returning query-relevant information at multiple granularities, but cannot reliably answer queries requiring complete-range filtering, aggregation, ranking, or temporal selection over append-only interaction histories.
    
    \item We propose \method, a multimodal memory framework that complements retrieval memory with schema-induced analytic memory. \method discovers recurrent structures from multimodal observations, materializes them as queryable tables, and uses a memory-aware planner to select or compose semantic retrieval with native analytical operations.
    
    \item We evaluate \method on two multimodal long-term memory benchmarks, MemEye and MemGallery, where \method consistently outperforms strong memory baselines, and improves accuracy by up to 11.3\% and 6.9\%, respectively.
\end{itemize}
\section{Related Work}
\paragraph{Agent memory.}
Long-term agent memory preserves information across interactions, allowing agents to reuse preferences, observations, decisions, and task states beyond a finite context window~\cite{needmemory,llmagents,ragsurvey}. Existing systems explore personalized stores and tiered memory management~\cite{MemoryBank,memgpt}, scalable or agentic memory construction~\cite{mem0,amem}, and temporal or relational organization~\cite{zep,licomemory}. Recent schema-grounded memory moves beyond semantic recall by transforming textual interactions into validated, queryable records, but relies on application-defined schemas~\cite{petrov2026xmemory}. Despite these advances, existing methods remain largely text-centric, leaving underexplored how to induce reusable analytical structures from heterogeneous multimodal histories and adaptively execute retrieval and analytical operations over them.

\paragraph{Multimodal agent memory.}
Multimodal agent memory extends text-centric memory to histories containing heterogeneous visual observations. 
Existing approaches broadly follow two directions: MIRIX and M2A translate visual content into captions, summaries, or semantic abstractions for text-based retrieval~\cite{mirix,m2a}, whereas MMA, VisualMem, and Omni-SimpleMem retain native visual evidence and retrieve multimodal or image-backed memories~\cite{mma,visualmem,omnisimplemem}. Although these methods improve visual evidence preservation, they still primarily organize histories as semantic memories. MemEye further reveals their difficulty in capturing fine-grained details and evolving visual states~\cite{memeye}. In contrast, \method adaptively organizes multimodal histories into complementary analytical and semantic memories, enabling both precise analysis and context-aware retrieval for complex queries.
\section{Method}

\begin{figure*}[t]
    \centering
    \includegraphics[width=\textwidth]{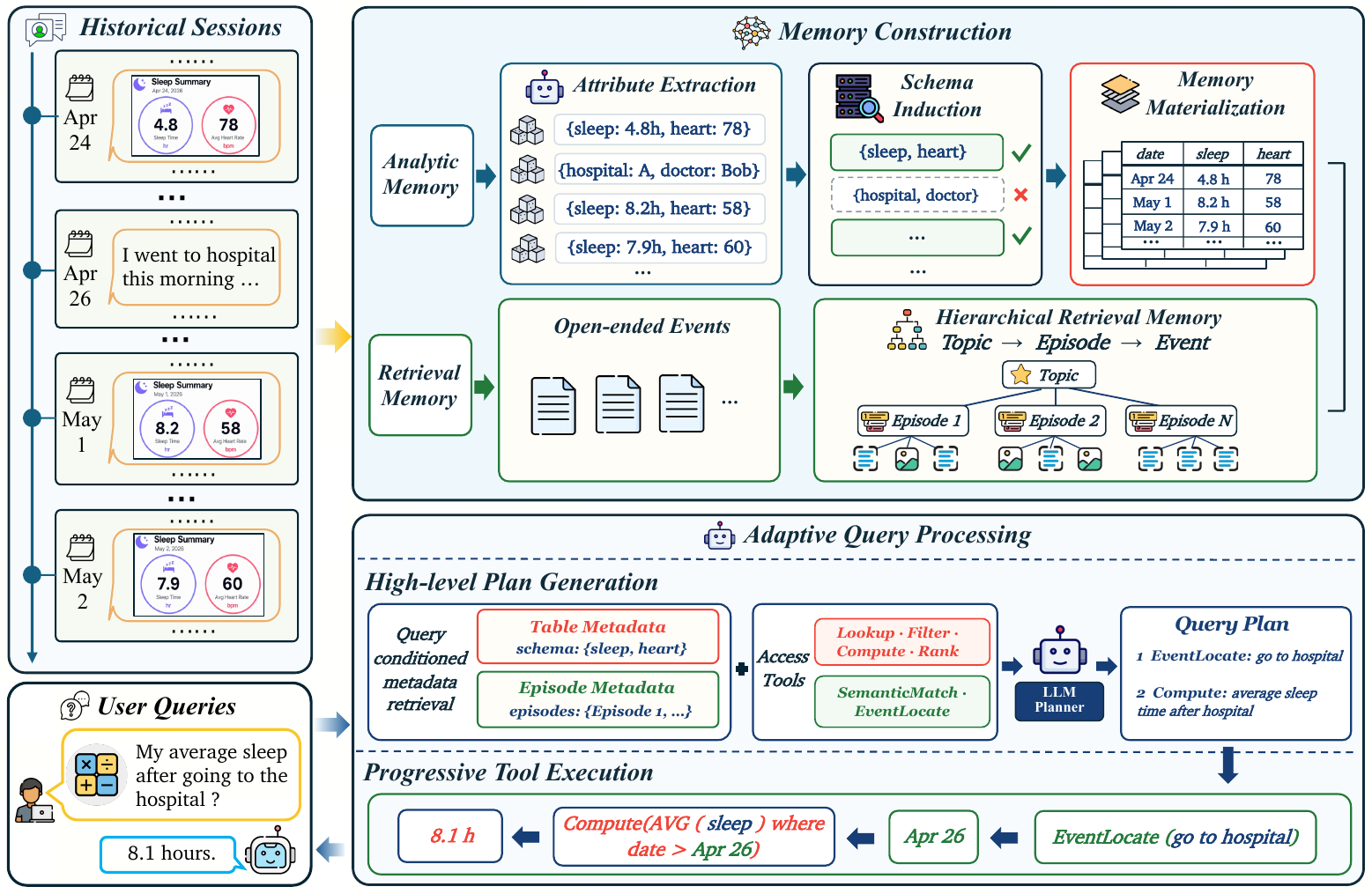}
    \caption{
    Overview of \method. It complements recall-oriented retrieval memory with schema-induced analytic memory over recurrent multimodal observations.
    At query time, an operation planner jointly considers the query and the instantiated memories to select designed access tools for answer generation.
    }
    \label{fig:overview_framework}
\end{figure*}

\subsection{Problem Formulation}

Let $\mathcal{H}=\{S_i\}_{i=1}^{M}$ denote a multimodal interaction history, where each session 
$S_i=\{R_{i,j}\}_{j=1}^{n_i}$ consists of multiple interaction rounds. 
Each round 
$R_{i,j}=(d_{i,j},\mathcal{V}_{i,j},\tau_{i,j})$ comprises the 
user--agent dialogue $d_{i,j}$, visual observations 
$\mathcal{V}_{i,j}$, and temporal information $\tau_{i,j}$. 
A multimodal agent memory system transforms the interaction history 
into memory representations,
$
\mathcal{M}=F_{\mathrm{build}}(\mathcal{H}),
$
and aims to generate an accurate answer $
\hat{y}=F_{\mathrm{answer}}(\mathcal{M},q)
$ for a user query $q$ based on $\mathcal{M}$.

\subsection{Overview}


Figure \ref{fig:overview_framework} presents the overview of the proposed \method framework.
Given a long-term multimodal interaction history, \method constructs two complementary memories: analytic memory $\mathcal{M}^{\mathrm{ana}}$  induces recurring patterns as adaptive schemas for analytical queries
while retrieval memory $\mathcal{M}^{\mathrm{ret}}$ hierarchically organizes interactions according to their semantics to enable flexible retrieval.
At query time, \method jointly reasons over the query and the memories instantiated from the current history, then constructs an execution plan by adaptively selecting and composing access tools to gather sufficient evidence for grounded answer generation.

\subsection{Analytic Memory Construction}
Analytic memory aims to uncover and organize recurring patterns across fragmented observations in multimodal interaction histories for analytical operations.
The key challenge is that the meaningful attributes, which is in the form of key-value pair, and their  co-occurrence patterns are unknown in advance, and each round reveals only a sparse fragment of the latent structure. 
To address this challenge, \method treats $\mathcal{H}$ as a chronologically ordered interaction rounds 
$\{R_t\}_{t=1}^{N}$, performs \emph{Attribute Extraction} to adaptively identify attribute evidence within each round, then conducts \emph{Schema Induction} to discover stable structures from recurring cross-round patterns, and finally applies \emph{Memory Materialization} to transform the induced schemas into executable analytic memory.



\subsubsection{Attribute Extraction}
Recovering recurring structure must begin from the evidence available within individual rounds, yet the attributes that will form a useful schema are unknown at extraction time. 
Record fragment extraction therefore aims to preserve explicit analytic observations from each round without imposing a predefined and unobserved pattern.

Given $R_t$, \method jointly examines its dialogue and visual content to identify every grounded attribute--value correspondence: 
\begin{equation}
\mathcal{O}_t
=
\operatorname{Extractor}(R_t)
=
\left\{
\left(
a_{t\ell},
x_{t\ell},
p_{t\ell}
\right)
\right\}_{\ell=1}^{m_t},
\label{eq:record-fragment-extraction}
\end{equation}
where $a_{t\ell}$ and $x_{t\ell}$ are the observed attribute and its source-faithful value, and $p_{t\ell}$ points to the supporting dialogue span or image.
$\operatorname{Extractor}$ is an LLM-based extractor that aims to identify all possible pairs from each round. 
The resulting collection $\{\mathcal{O}_t\}_{t=1}^{N}$  captures these observations and their within-interaction co-occurrence patterns for subsequent schema induction.

\subsubsection{Schema Induction}

Given the extracted record fragments, it remains challenging to induce reusable schemas without assuming a predefined schema.
New observations may reveal either a previously unseen structure or evolution to an existing one. 
To address this challenge, \method first mine candidate patterns and then induces schemas through two processes: 
\emph{Schema Discovery} creates new schemas, while \emph{Schema Evolution} extends existing ones with consistently co-occurring attributes.

\paragraph{Candidate Pattern Mining.}
For each extracted record fragment $\mathcal{O}_t$, let
\begin{equation}
\mathcal{A}_t
=
\left\{
a_{t\ell}
\mid
\left(
a_{t\ell},
x_{t\ell},
p_{t\ell}
\right)
\in
\mathcal{O}_t
\right\}
\label{eq:attribute-set}
\end{equation}
denote the set of attributes contained in $\mathcal{O}_t$.
A reliable candidate pattern $C$ is a set of attributes that  consistently recur sufficient interactions together.
Following the Apriori framework for frequent-itemset
mining~\citep{agrawal1994fast}, we use \emph{support} to quantify information recurrence. 
Specifically, the \emph{support} of a candidate pattern $\mathcal{C}$ after processing round $t$ is
\begin{equation}
    \operatorname{supp}_t(\mathcal{C})
    =
        \sum_{s=1}^{t}
        \mathbb{I}\!\left[C\subseteq \mathcal{A}_s\right],
\end{equation}

\method retains $\mathcal{C}$ as a candidate pattern if
\begin{equation}
    \operatorname{supp}_t(\mathcal{C})\geq\theta_s,
\end{equation}
which ensures that $\mathcal{C}$ recurs across sufficient
interaction rounds to yield a populated schema. 

\paragraph{Schema Discovery.} 
For each recurrent candidate $\mathcal{C}$, \method first compares it with the active schemas before round $t$. 
If $\mathcal{C}$ exhibits an attribute structure distinct from all existing schemas, it is evaluated for reliable new-schema discovery.
Following a variant of Apriori algorithm~\citep{omiecinski2003alternative}, we use \emph{all-confidence} to measure within-pattern co-occurrence consistency, defined as
\begin{equation}
    \operatorname{all-conf}_t(\mathcal{C})
    =
    \frac{
        \operatorname{supp}_t(\mathcal{C})
    }{
        \max_{a\in \mathcal{C}}\operatorname{supp}_t(\{a\})
    }.
\end{equation}
A candidate is admitted as a new schema if
\begin{equation}
    \operatorname{all-conf}_t(\mathcal{C})\geq\theta_a.
\end{equation}
This constraint requires the complete candidate to occur reliably whenever any of its constituent attributes occurs, thereby preventing an ubiquitous attribute from being merged with most patterns. 
To avoid admitting a qualified candidate together with its qualified subsets, \method retains only
inclusion-maximal novel candidates.

\paragraph{Schema Evolution.}
For an existing schema $\mathcal{A}$, if $\mathcal{C}\subseteq \mathcal{A}$, then $\mathcal{C}$ is already subsumed by it.
If $\mathcal{C}$ contains $\mathcal{A}$, \method treats $\mathcal{C}$ as a potential extension of $\mathcal{A}$. 
Let $ \Delta=\mathcal{C}\setminus \mathcal{A}$ be the newly added attributes. 
We use \emph{extension confidence} to measure how consistently $\Delta$ accompany occurrences of $\mathcal{A}$, defined as
\begin{equation}
    \operatorname{ext-conf}_t(\mathcal{C} \mid \mathcal{A})
    =
    \frac{
        \operatorname{supp}_t(\mathcal{C})
    }{
        \operatorname{supp}_t(\mathcal{A})
    } 
\end{equation}
When $\operatorname{ext-conf}_t(\mathcal{C}\mid \mathcal{A})\geq\theta_e$, the candidate provides sufficient evidence that $\Delta$ have become stable components of $\mathcal{A}$, and we update the schema $\mathcal{A}$ to $\mathcal{C}$. Otherwise, $\mathcal{A}$ remains unchanged.

\subsubsection{Memory Materialization}
Considering that each schema specifies a stable set of attributes and each matched interaction provides their co-occurring values, the induced patterns naturally align with relational tables. 
\method therefore materializes each schema $\mathcal{A}$ into a table $T_A$, using its attributes as columns and the matched interactions as rows.


Specifically, for a newly discovered schema $\mathcal{A}$, its attributes first define the data columns.
Given these attributes and their sampled values from the corresponding extracted observations $\{\mathcal{O}_t\}_{t=1}^{N}$, a language model generates the table metadata, including its name and description, and infers the type of each column.
We then populate the table at the interaction level.
For each associated round $R_t$, every extracted tuple $(a_{t\ell},x_{t\ell},p_{t\ell})\in\mathcal{O}_t$ whose attribute belongs to $\mathcal{A}$ is mapped to the corresponding column, and values extracted from the same round jointly form one row.
Two auxiliary columns, \texttt{order} and \texttt{time}, record the position and occurrence time of the source interaction, respectively.
Any value not observed in that round is left empty.
When a schema evolves from $\mathcal{A}$ to
$\mathcal{A}'=\mathcal{A}\cup\Delta$, the attributes in $\Delta$ are appended as new columns to the existing table $T_{\mathcal{A}}$.
\method then updates the table information and rematerializes the rows from their associated observations following the same procedure.
The resulting analytic memory $\mathcal{M}^{\mathrm{ana}}$ is represented as the collection of tables $T_{\mathcal{A}}$.

\subsection{Retrieval Memory Construction}

Beyond structured analytic queries, many user requests can be addressed through direct semantic matching against past interactions. Retrieval memory is designed for such queries by preserving open-ended events, relations, and visual details that may not exhibit recurring schemas. It complements analytic memory by enabling flexible semantic retrieval and context-aware reasoning over interaction histories.

Following prior work on hierarchical memory organization~\citep{yue2026hypermem}, we organize retrieval memory $\mathcal{M}^{\mathrm{ret}}$ into three levels: $\text{topic} \rightarrow \text{episode} \rightarrow \text{event}$.
Events preserve fine-grained evidence within individual interactions, episodes group temporally adjacent and semantically coherent events, and topics aggregate related episodes into broader semantic contexts. 
Each level maintains a concise description, separate textual and visual representations, and temporal information.
Appendix~\ref{app:hierarchical-retrieval-memory} details its incremental construction.

\subsection{Adaptive Query Processing}
\label{sec:adaptive-query-processing}


Different queries place different demands on memory: some require semantic retrieval over open-ended events, whereas others require structured computation over analytic memory. No single retrieval primitive can adequately support this full range of needs, calling for an adaptive mechanism that selects and composes operations according to query intent and the structures available in memory.
\method therefore abstracts the native capabilities of analytic and retrieval memory as operation-specific tools, and employs a joint query--memory planner to generate an executable procedure for answering each query.


\subsubsection{Memory Access Tools}
\label{sec:memory-access-tools}


Analytic and retrieval memory differ in their underlying structures and consequently provide distinct native capabilities.
To expose these capabilities at query time, \method equips each memory with structure-specific tools.


Specifically, each tool is represented as 
\begin{equation}
    \xi_i=(d_i,\Theta_i,\Omega_i),
    \label{eq:memory-interface}
\end{equation}
where $d_i$ textually specifies its tool function, while $\Theta_i$ and $\Omega_i$ define its typed arguments and return schema.
Table~\ref{tab:core-memory-access-interfaces} summarizes the access tools designed for analytic and retrieval memory.
For example, for the analytic tool \textsc{Compute}, $d_i$ specifies a deterministic computation over an induced table, $\Theta_i$ includes the target table, computation operator, target column, and optional constraints, and $\Omega_i$ returns the computed result and matched records. 
For the retrieval tool \textsc{SemanticMatch}, $d_i$ specifies relevance-based memory retrieval, $\Theta_i$ includes semantic query content and a retrieval budget, and $\Omega_i$ returns ranked memory units with their relevance scores. Detailed specifications are provided in Appendix~\ref{app:tool-specifications}.

\begin{table}[t]
\centering
\small
\caption{Access tools for analytic and retrieval memory.}
\label{tab:core-memory-access-interfaces}
\renewcommand{\arraystretch}{1.08}
\begin{tabularx}{\columnwidth}{
    >{\raggedright\arraybackslash}p{0.17\columnwidth}
    >{\raggedright\arraybackslash}p{0.36\columnwidth}
    >{\raggedright\arraybackslash}X
}
\toprule
\textbf{Memory} & \textbf{Tools} & \textbf{Capability} \\
\midrule
Analytic
& \textsc{Lookup}, \textsc{Filter}, \textsc{Compute}, \textsc{Rank}
& Exact record access and deterministic computation. \\

Retrieval
& \textsc{SemanticMatch}, \textsc{EventLocate}
& Relevance-based retrieval and event localization. \\
\bottomrule
\end{tabularx}
\end{table}






\subsubsection{Memory-Aware Joint Query Planning}

The access tools specify how each memory can be queried.
However, users describe desired outcomes rather than access procedures, and a feasible operation also depends on what the current memory contains.
To bridge this gap, \method first constructs a query-conditioned planning context that exposes relevant 
memory structure, and then employs an LLM-based planner to jointly reason over the query and this context, composing an executable procedure for retrieving and analyzing the required information. 
The resulting evidence is subsequently provided to the answer model for response generation.




\paragraph{Planning Context Construction.}
Valid query planning requires knowledge of the current memory structures and their supported operations. 
Exposing the complete memory contents, however, introduces irrelevant information and expands the planning space. 
\method therefore abstracts analytic and retrieval memory into compact metadata, retrieves query-relevant entries, and associates them with their available tools to form the effective planning context.

Specifically, we treat each analytic table $T\in\mathcal{M}^{\mathrm{ana}}$ and retrieval episode $P\in\mathcal{M}^{\mathrm{ret}}$ as an individual planning candidate represented by textual metadata $\mu_T$ and $\mu_P$, respectively. For each table, $\mu_T$ summarizes its name, description, columns, and sampled values, while each episode directly uses its description as $\mu_P$. Together, these candidates form the metadata search space $\mathcal{U}$.
Given a query $q$, \method ranks each $\mu\in\mathcal{U}$ using
\begin{equation}
    \begin{aligned}
        h_{\lambda}(q,\mu)
        &=
        \lambda\cos\!\left(E(q),E(\mu)\right) \\
        &\quad+
        (1-\lambda)\operatorname{TokSim}(q,\mu),
    \end{aligned}
    \label{eq:metadata-relevance}
\end{equation}
where $E(\cdot)$ is a semantic encoder and $\operatorname{TokSim}(\cdot)$ is computed using BM25 over the tokenized query and metadata, capturing lexical matches following prior works \cite{m2a}.
After retaining the top-ranked $\mu$, \method augments their metadata with designed access tool descriptions, forming the planning context $\mathcal{D}_q$ to expose relevant memory candidates and available operations to the planner.

\paragraph{Progressive Tool-Execution Planning.}
Complex queries often require multiple dependent operations, with the arguments of later tool calls determined by earlier outputs.
Instantiating all tool calls upfront is therefore unreliable. To address this challenge, \method separates high-level planning from progressive tool instantiation.

\emph{High-level plan generation.}
Conditioned on the query $q$ and planning context $\mathcal{D}_q$, \method first employs an LLM-based planner $\operatorname{Planner}(\cdot)$ to generate an abstract plan
$\bar{\pi}(q, \mathcal{D}_q) = \operatorname{Planner}(q,\mathcal{D}_q) =
    \big[(g_\ell,\xi_\ell)\big]_{\ell=1}^{L}$, which specifies the information goals $g_\ell$ and the selected tool $\xi_\ell$ at step $\ell$ without binding them to concrete arguments.

\emph{Progressive tool instantiation.}
Guided by the above high-level plan, \method instantiates tool calls sequentially.
At step $\ell$, it conditions on the preceding results 
$\mathcal{Z}_{<\ell}$ and optimizes:
\begin{equation}
    \begin{aligned}
        &\Theta_\ell
         =
        \operatorname{Planner}
        (q,\mathcal{D}_q,\mathcal{Z}_{<\ell}, g_\ell,\xi_\ell),\\
        &\mathcal{Z}_{<\ell+1} = \mathcal{Z}_{<\ell} 
        \cup \operatorname{Invoke}(\xi_\ell,\Theta_\ell),
    \end{aligned}
    \label{eq:progressive-tool-instantiation}
\end{equation}
where $\xi_\ell$ and $\Theta_\ell$ denote the selected tool and its instantiated arguments.
This allows later calls to consume intermediate results (e.g.,  a timestamp returned by \textsc{EventLocate} can serve as a temporal constraint for a subsequent \textsc{Compute} call).

\paragraph{Question Answering.}
Finally, the answer model generates response conditioned on the query and an evidence context comprising the instantiated plan and its tool outputs.

\section{Experimental Evaluation}


\subsection{Experimental Setup}

\paragraph{Benchmarks.}
We evaluate \method on two complex multimodal memory benchmarks. 
\textbf{MemEye} evaluates visual-memory granularity and reasoning through paired multiple-choice and open-ended queries~\citep{memeye}, while \textbf{MemGallery} assesses long-term conversational memory management~\citep{memgallery}. 
\textbf{MemEye} uses EM, BLEU-1 and LLM-judge for evaluation, 
whereas \textbf{MemGallery} reports F1, BLEU-1, and LLM-Judge scores.
Dataset statistics are provided in Table~\ref{tab:dataset_statistics}.

\begin{table}[t]
\centering
\small
\begin{tabular}{lrrrr}
\toprule
Benchmarks & Sessions & Rounds & Images & QA pairs  \\
\midrule
MemEye & 221 & 848 & 438 & 742 \\
MemGallery & 240 & 3,962 & 1,003 & 1,711 \\
\bottomrule
\end{tabular}
\caption{Benchmark statistics.}
\label{tab:dataset_statistics}
\end{table}

\paragraph{Baselines.}
We compare against unimodal memory agents, including \textbf{A-Mem} and \textbf{MemoryOS}~\citep{amem,memos}, which organize long-term textual interaction histories. 
Multimodal baselines include dedicated memory agents (\textbf{M2A}, \textbf{MMA}, and \textbf{MIRIX})~\citep{m2a,mma,mirix}, which construct persistent memories from dialogue and visual observations, and retrieval-based systems (\textbf{MM-RAG} and \textbf{UniversalRAG}), which directly retrieve relevant multimodal evidence for answer generation.


\paragraph{Implementation details.}
We use GPT-4.1-nano and GPT-5.4-mini as answer and memory construction backbones for all methods.
Text and image representations are produced by MiniLM-L6-v2~\citep{wang2020minilm} and siglip2-base-patch16-384~\citep{tschannen2025siglip2}, respectively.
We use the top-10 retrieved memory units for retrieval memory baselines, and constrain our planner to generate at most three execution steps under a shared budget of 10 evidence units.
Results are averaged over 3 runs. 
Complete parameter and prompt details are provided in Appendix~\ref{app:implementation-details} and \ref{app:prompt-templates}.

\subsection{Main Results}

\begin{table*}[t]
\centering
\small
\setlength{\tabcolsep}{2pt}
\begin{tabular*}{\textwidth}{@{\extracolsep{\fill}}lllcccccc@{}}
\toprule
\multirow{2}{*}{Backbone}
& \multicolumn{2}{c}{\multirow{2}{*}{Method}}
& \multicolumn{3}{c}{MemEye}
& \multicolumn{3}{c}{MemGallery} \\
\cmidrule(lr){4-6}\cmidrule(lr){7-9}
& \multicolumn{2}{c}{}
& EM & BLEU-1 & LLM-Judge & F1 & BLEU-1 & LLM-Judge \\
\midrule
\multirow{8}{*}{GPT-4.1-nano}
& \multirow{2}{*}{Unimodal}
& A-Mem      & 39.1 & 17.1 & 32.0 & 53.2 & 47.2 & 67.4 \\
& & MemoryOS & 42.4 & 15.8 & 27.6 & 53.4 & 47.1 & 65.9 \\
\cmidrule(lr){2-3}
& \multirow{6}{*}{Multimodal}
& M2A          & 34.1 & 5.8 & 13.8 & 50.2 & 44.3 & 61.5 \\
& & MMA        & 39.0 & 10.6 & 32.0 & 55.9 & 50.7 & 64.1 \\
& & MIRIX      & 40.9 & 5.7 & 13.6 & 56.3 & 51.1 & 66.3 \\
& & MM-RAG     & 42.7 & 16.6 & 42.2 & 58.3 & 52.4 & 67.3 \\
& & UniversalRAG
               & 43.1 & 16.5 & 40.2 & 57.7 & 52.4 & 67.0 \\
& & \method \textbf{(Ours)}
& \textbf{50.4}\textcolor{red}{\scriptsize(+7.3)}
& \textbf{21.2}\textcolor{red}{\scriptsize(+4.1)}
& \textbf{48.0}\textcolor{red}{\scriptsize(+5.8)}
& \textbf{62.6}\textcolor{red}{\scriptsize(+4.3)}
& \textbf{56.7}\textcolor{red}{\scriptsize(+4.3)}
& \textbf{74.3}\textcolor{red}{\scriptsize(+6.9)} \\
\midrule
\multirow{8}{*}{GPT-5.4-mini}
& \multirow{2}{*}{Unimodal}
& A-Mem      & 48.0 & 22.4 & 35.2 & 63.0 & 58.6 & 74.4 \\
& & MemoryOS & 48.7 & 21.8 & 33.3 & 64.5 & 59.0 & 75.5 \\
\cmidrule(lr){2-3}
& \multirow{6}{*}{Multimodal}
& M2A          & 40.1 & 12.3 & 33.5 & 60.8 & 56.0 & 71.7 \\
& & MMA        & 53.9 & 25.7 & 43.3 & 64.0 & 59.6 & 73.3 \\
& & MIRIX      & 46.7 & 18.0 & 33.2 & 65.7 & 60.9 & 78.2 \\
& & MM-RAG     & 61.8 & 29.7 & 49.4 & 66.5 & 61.9 & 78.7 \\
& & UniversalRAG
               & 62.4 & 30.4 & 48.6 & 65.6 & 60.8 & 77.4 \\
& & \method \textbf{(Ours)}
& \textbf{65.5}\textcolor{red}{\scriptsize(+3.1)}
& \textbf{35.9}\textcolor{red}{\scriptsize(+5.5)}
& \textbf{60.7}\textcolor{red}{\scriptsize(+11.3)}
& \textbf{69.1}\textcolor{red}{\scriptsize(+2.6)}
& \textbf{64.1}\textcolor{red}{\scriptsize(+2.2)}
& \textbf{83.9}\textcolor{red}{\scriptsize(+5.2)} \\
\bottomrule
\end{tabular*}
\caption{Main results on MemEye and MemGallery. Higher is better for all metrics. \textbf{Bold values} denote the best result within a backbone and \textcolor{red}{red values} denote improvements over the strongest baseline.}
\label{tab:main_results}
\end{table*}

\paragraph{Overall performance.}
Table~\ref{tab:main_results} presents a comprehensive comparison of \method with representative unimodal and multimodal memory frameworks across two benchmarks and answer backbones.
\method consistently achieves the best result on every metric.
With GPT-4.1-nano, it surpasses the strongest competing result on MemEye by 7.3\% and 5.8\% percentage points for MCQ and open-ended questions, respectively, and improves MemGallery F1, BLEU-1, and LLM-Judge by 4.3\%, 4.3\%, and 6.9\%.
The advantage persists with GPT-5.4-mini, yielding gains of 3.1\% and 11.3\% points on MemEye and 2.6\%, 2.2\%, and 5.2\% on the three MemGallery metrics.
The consistent improvements on different settings demonstrate that \method generalizes across answer formats, benchmark settings, and backbone capacities.

\begin{figure}[t]
\centering
\includegraphics[width=\linewidth]{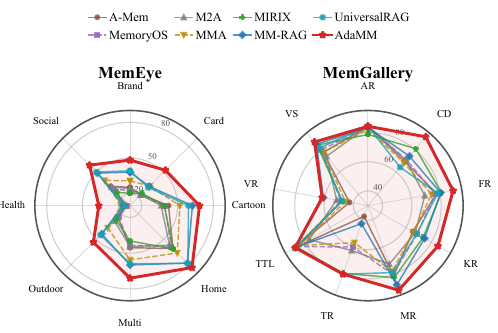}
\caption{LLM-as-Judge performance across fine-grained  tasks with GPT-5.4-mini.}
\label{fig:task_correctness}
\end{figure}

\paragraph{Fine-grained task analysis.}
To further examine where the improvement arises, Figure~\ref{fig:task_correctness} reports the LLM-Judge breakdown using GPT-5.4-mini.
On MemEye, the largest gains over the strongest baseline occur on Card Playlog and Personal Health, with improvements of 18.8\% and 16.7\%, respectively.
Both tasks require exact operations over recurring records, such as filtering observations, comparing values, or tracking changes, highlighting the benefit of analytic memory.
\method also achieves substantial gains on Multi-scene (11.1\%), Brand Memory (9.0\%), Outdoor Navigation (8.6\%), and Social Chat (8.2\%), where relevant evidence must be retrieved and connected across interactions.
The smaller gains on Cartoon Entertainment (6.1\%) and Home Renovation (5.2\%) are consistent with their greater reliance on direct visual-semantic recall.
On MemGallery, the largest margins appear in Conflict Detection (10.6\%), Knowledge Resolution (10.5\%), and Factual Retrieval (8.1\%), demonstrating the value of structured access for reconciling and precisely locating information.
\method further improves Multi-entity Reasoning, Visual-centric Reasoning, and Visual-centric Search, while remaining comparable to the strongest baselines on Answer Refusal, Temporal Reasoning, and Test-time Learning.
Together, these results suggest that combining retrieval with operation-specific analytic access is particularly beneficial when answering requires more than semantic relevance alone.
Further analysis of cost and case study are provided in Appendix \ref{app:cost-analysis} and \ref{app:case-study}.

\subsection{Ablation Study}

\begin{figure}[t]
\centering
\includegraphics[width=\linewidth]{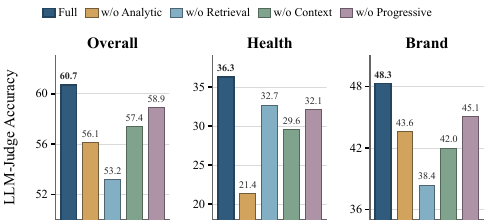}
\caption{\textbf{Ablation results on MemEye.}
Overall and task-level LLM-as-Judge accuracy using GPT-5.4-mini.}
\label{fig:structure_ablation}
\end{figure}

We evaluate four ablated variants on MemEye to examine the contributions of the designed memories and the memory-aware planning mechanism.
The Health task emphasizes numerical and temporal analysis over multiple personal-health records, whereas the Brand task primarily evaluates visually grounded recall of brand-related information.
In \emph{w/o Analytic} and \emph{w/o Retrieval}, we disable the analytic memory and retrieval memory, respectively.
In \emph{w/o Planning Context}, the planner receives the entire memory metadata.
In \emph{w/o Progressive Execution}, all tool calls and their arguments are instantiated before execution.

\paragraph{Contributions of Complementary Memories.}
As shown in Figure~\ref{fig:structure_ablation}, removing either memory component consistently degrades performance.
Without analytic memory, overall accuracy decreases by 4.6\%, with a substantially larger 14.9\% drop on Health.
This result highlights the importance of organizing recurring records into executable structures for analytical queries.
Removing retrieval memory instead causes a 7.5\% overall drop and is particularly detrimental to Brand, where performance decreases by 9.9\%.
The distinct degradation patterns confirm that analytic and retrieval memory provide complementary capabilities.

\paragraph{Contributions of Memory-Aware Planning.}
Removing the planning context reduces performance by 3.3\% overall and by 6.7\% and 6.3\% on Health and Brand, respectively.
This demonstrates that effective tool selection requires knowledge of both the query intent and the memory structures currently available.
Instantiating all calls before execution also produces consistent degradation, including drops of 4.2\% on Health and 3.2\% on Brand.
These results indicate that progressive execution is useful when later operations depend on information obtained from preceding calls.




\section{Conclusion}

In this work, we identify a retrieval--analysis mismatch in long-term multimodal agent memory, highlighting the need for executable operations over accumulated experience beyond retrieval.
To address this issue, we introduced \method, which couples hierarchical retrieval memory with schema-induced analytic memory and exposes their distinct capabilities through operation-specific tools.
A memory-aware planner further grounds tool selection in the current memory state and progressively composes retrieval and analytic operations.
Experiments on MemEye and MemGallery across two answer backbones demonstrate consistent improvements from combining flexible recall with executable analysis.

\section*{Limitations}


While \method demonstrates promising results in long-term multimodal agent memory, two limitations remain.
First, analytic memory depends on record fragments extracted from multimodal interactions. Incorrect or missing fields may propagate to schema induction, table construction, and downstream computations. Future work could incorporate confidence-aware extraction and cross-round consistency checks to improve robustness.
Second, \method uses a predefined set of access tools and thus requires manual extension for unseen, domain-specific operations. Future work could
develop a self-evolving tool framework that identifies emerging capability gaps and safely synthesizes, validates, and integrates new tools.

\bibliography{custom}
\appendix

\section{Hierarchical Retrieval Memory Construction}
\label{app:hierarchical-retrieval-memory}

Following the hierarchical organization commonly adopted in text-based agent memory~\citep{yue2026hypermem}, we structure multimodal interaction histories into three levels---topic, episode, and event---enabling semantic retrieval to progressively localize relevant memories from coarse-grained themes to fine-grained interactions.

Given chronologically ordered rounds $\{R_t\}_{t=1}^{N}$, \method incrementally inserts new events, updates the current episode, and assigns completed episodes to topics.

\paragraph{Event construction.}
Each interaction round $R_t$ is directly instantiated as an event. The event retains the round's dialogue and contextual content, interaction order, timestamp, and associated image identifiers. Thus, event construction preserves the original round-level evidence without an additional language-model extraction step.

\paragraph{Multimodal representations.}
Each hierarchy node $v$ maintains separate textual and visual embeddings. Its textual representation is
\begin{equation}
    \mathbf{z}^{\mathrm{text}}_v
    = \operatorname{Encoder}_{\mathrm{text}}(d_v),
    \label{eq:hierarchical-text-embedding}
\end{equation}
where $d_v$ is the textual content of an event or the current episode/topic description. For an event $e$ with associated images $\mathcal{X}_e$, its visual representation is obtained by mean pooling their visual embeddings and then applying $\ell_2$ normalization:
\begin{equation}
    \mathbf{z}^{\mathrm{vis}}_e
    = \operatorname{Norm}\!\left(
        \frac{1}{|\mathcal{X}_e|}
        \sum_{x\in\mathcal{X}_e}
        \operatorname{Encoder}_{\mathrm{visual}}(x)
      \right).
    \label{eq:event-visual-embedding}
\end{equation}
The visual representation of an episode or topic is computed analogously by pooling only the visual embeddings of children that contain images. Textual and visual embeddings are not projected into one space; their cosine similarities are combined through late fusion,
\begin{equation}
    \begin{aligned}
        s(u,v)
        ={}& \bar{\lambda}_{\mathrm{text}}
        \cos(\mathbf{z}^{\mathrm{text}}_u,
             \mathbf{z}^{\mathrm{text}}_v) \\
        &+ \bar{\lambda}_{\mathrm{vis}}
        \cos(\mathbf{z}^{\mathrm{vis}}_u,
             \mathbf{z}^{\mathrm{vis}}_v),
    \end{aligned}
    \label{eq:hierarchical-multimodal-similarity}
\end{equation}
where the weights are normalized over the modalities available to both nodes. Thus, the visual term is omitted rather than replaced by a zero vector when either node has no visual evidence. During retrieval, a textual query is encoded by $\operatorname{Encoder}_{\mathrm{text}}$ for text-to-text matching and by the text branch of $\operatorname{Encoder}_{\mathrm{visual}}$ for text-to-image matching, following the same fusion rule.

\paragraph{Episode construction.}
Events are processed in temporal order. A new event is compared only with the current episode using Eq.~\ref{eq:hierarchical-multimodal-similarity}. It is appended when the temporal gap is within one day and the multimodal similarity exceeds $\theta_{\mathrm{epi}}$; otherwise, the current episode is closed and a new one is initialized. After insertion, a language model regenerates the episode description from the ordered contents of all its constituent events. Its textual embedding is recomputed from the updated description, while its visual embedding is recomputed from the image-bearing child events.

\paragraph{Topic construction.}
When an episode is closed, it is compared with all existing topics without imposing temporal adjacency. The episode is attached to its highest-scoring topic if the fused similarity exceeds $\theta_{\mathrm{topic}}$; otherwise, a new topic is created. A language model generates the topic description from all episode descriptions currently assigned to it. The textual and visual embeddings are then updated accordingly. Consequently, each new interaction only updates its event, episode, and topic ancestors, enabling the hierarchy to evolve without reconstructing the complete memory.


\section{Memory Access Tool Specifications}
\label{app:tool-specifications}

The six access tools expose the native operations of analytic and retrieval memory through a common typed interface. 
Each tool is implemented as a concrete programmatic function with a predefined execution procedure; the planner only selects the tool and instantiates its arguments. 
Table~\ref{tab:memory-access-tools} specifies the invocation arguments $\Theta_i$, returned results $\Omega_i$, and function of each tool $\xi_i$; the typewriter-formatted entries are the exact keys exposed to the planner.
For tools that may return multiple records or memory units, the result budget bounds the evidence passed to subsequent planning steps and answer generation.

\begin{table*}[t]
\centering
\small
\caption{Detailed access-tool specifications for analytic and retrieval memory.}
\label{tab:memory-access-tools}
\renewcommand{\arraystretch}{1.15}
\begin{tabularx}{\textwidth}{
    >{\raggedright\arraybackslash}p{0.10\textwidth}
    >{\raggedright\arraybackslash}p{0.16\textwidth}
    >{\raggedright\arraybackslash}p{0.24\textwidth}
    >{\raggedright\arraybackslash}p{0.20\textwidth}
    >{\raggedright\arraybackslash}X
}
\toprule
\textbf{Memory} &
\textbf{Tool} &
\textbf{Arguments $\Theta_i$} &
\textbf{Return $\Omega_i$} &
\textbf{Function} \\
\midrule

\multirow{4}{*}{Analytic}
& \textsc{Lookup}
& \texttt{table}; \texttt{key\_values}; \texttt{result\_budget}
& \texttt{matching\_rows}
& Retrieves records matching the specified keys. \\

& \textsc{Filter}
& \texttt{table}; \texttt{column\_filters}; \texttt{result\_budget}
& \texttt{matching\_rows}
& Selects records satisfying the specified predicates. \\

& \textsc{Compute}
& \texttt{table}; \texttt{target\_column\_expression}; \texttt{operator}; \texttt{optional\_column\_filters}
& \texttt{computed\_result}; \texttt{contributing\_rows}
& Performs deterministic computation over a selected table scope. \\

& \textsc{Rank}
& \texttt{table}; \texttt{target\_column\_expression}; \texttt{order}; \texttt{optional\_column\_filters}; \texttt{result\_budget}
& \texttt{ranked\_rows}
& Returns the top- or bottom-ranked records. \\

\midrule

\multirow{2}{*}{Retrieval}
& \textsc{SemanticMatch}
& \texttt{query\_content}; \texttt{retrieval\_scope}; \texttt{result\_budget}
& \texttt{ranked\_memory\_units}; \texttt{relevance\_scores}
& Retrieves semantically relevant memory units. \\

& \textsc{EventLocate}
& \texttt{query\_content}; \texttt{result\_budget}
& \texttt{ranked\_memory\_units}; \texttt{relevance\_scores}
& Locates event occurrences and their surrounding context. \\

\bottomrule
\end{tabularx}
\end{table*}

\paragraph{Analytic tool execution.}
Analytic tools execute read-only operations over a selected induced table. 
Before execution, table and column names are resolved against its metadata, supplied values are cast to the corresponding column types, and operator--type compatibility is validated. 
\textsc{Lookup} matches the specified target column--value pairs, whereas 
\textsc{Filter} selects rows satisfying the supplied column filters. 
\textsc{Compute} applies the designated operator to a target column expression over the optionally filtered rows and returns both the computed result and the matched rows. 
\textsc{Rank} evaluates the target column expression, orders the eligible rows as requested, and returns the highest- or lowest-ranked rows within the result budget.

\paragraph{Retrieval tool execution.}
Retrieval tools operate over the topic--episode--event hierarchy. 
\textsc{SemanticMatch} scores memory units against the query content using the multimodal similarity in Eq.~\ref{eq:hierarchical-multimodal-similarity}, restricts candidates to the requested hierarchy scope, and returns the highest-scoring units within the result budget. \textsc{EventLocate} specializes this process for event occurrences: it identifies query-relevant events, augments each with its fixed window of surrounding context, and returns the ranked event-centered memory units. 
Both tools expose the relevance score of each returned memory unit and corresponding meta information (e.g., time, round order, etc.) for downstream planning and answering.




\FloatBarrier

\section{Implementation Details}
\label{app:implementation-details}

\paragraph{Models and decoding.}
We evaluate \method with GPT-4.1-nano and GPT-5.4-mini as two separate backbone settings. Within each setting, the same language model performs attribute extraction, table-metadata and hierarchy-description generation, query planning, and final answer generation. We use deterministic decoding with temperature 0. Textual descriptions and metadata are encoded with \texttt{MiniLM-L6-v2}~\citep{wang2020minilm}, while images are encoded with \texttt{siglip2-base-patch16-384}~\citep{tschannen2025siglip2}; the latter's text branch embeds textual queries for text-to-image matching. 

\paragraph{Parameter settings.}
Table~\ref{tab:implementation-hyperparameters} lists all manually specified parameters used by \method. 
We select these settings on the validation split and keep them fixed across both benchmarks and backbone models. 
The schema support threshold is an absolute interaction count. 
When visual evidence is unavailable, the text weight is renormalized to one.

\begin{table}[H]
\centering
\small
\caption{Key hyperparameter settings.}
\label{tab:implementation-hyperparameters}
\renewcommand{\arraystretch}{1.08}
\begin{tabularx}{\columnwidth}{
    >{\raggedright\arraybackslash}X
    >{\raggedleft\arraybackslash}p{0.28\columnwidth}
}
\toprule
\textbf{Parameter} & \textbf{Setting} \\
\midrule
Schema support threshold & $\theta_s=4$ rounds \\
Schema all-confidence threshold & $\theta_a=0.60$ \\
Schema extension-confidence threshold & $\theta_e=0.80$ \\
Episode similarity threshold & $\theta_{\mathrm{epi}}=0.60$ \\
Topic similarity threshold & $\theta_{\mathrm{topic}}=0.65$ \\
Text/visual fusion weights & $0.60/0.40$ \\
Metadata semantic/token weights & $0.70/0.30$ \\
Retrieved table/episode metadata & $k_{\mathrm{tab}}=k_{\mathrm{epi}}=5$ \\
Maximum planning steps & 3 \\
Shared evidence budget & 10 units \\
\bottomrule
\end{tabularx}
\end{table}

For budgeting, one retrieved memory unit in retrieval memory or one matched row in analytic memory is viewed as one evidence unit; a computed result and its contributing records are packaged as one unit. 
No tool call may exceed the remaining shared budget. 
Retrieval-based baselines use the same total allowance through top-10 retrieval. We report averages over three runs.

\FloatBarrier

\section{Cost Analysis}
\label{app:cost-analysis}

Following prior work, we use token consumption to measure the cost of query processing over all test queries under the same backbone and evidence budget.
For each method, we count the total input and output tokens consumed by all language-model calls triggered by a query, including planning, progressive tool instantiation, and final answer generation.
For multimodal calls, this total includes the image-input tokens reported by the model provider.

\begin{figure}[t]
\centering
\includegraphics[width=\columnwidth]{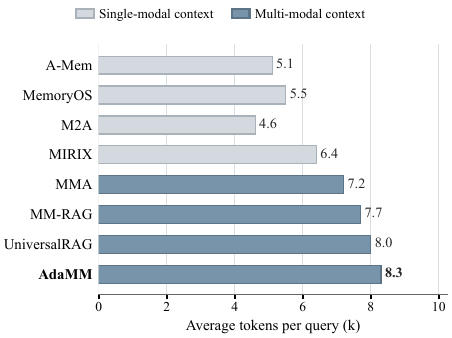}
\caption{\textbf{Query-time token consumption on MemEye.}
Average input and output tokens consumed per query using GPT-5.4-mini. Lower is better.}
\label{fig:query-token-cost}
\end{figure}

Figure~\ref{fig:query-token-cost} shows that query-time token consumption largely depends on whether original images are included in the answer context.
A-Mem, MemoryOS, M2A, and MIRIX answer from textualized memories and consume 4.6--6.4k tokens per query, whereas MMA, MM-RAG, UniversalRAG, and \method retain visual evidence and incur higher token costs.
\method consumes 8.3k tokens, only 0.6k more than MM-RAG and 0.3k more than UniversalRAG.
This modest overhead arises from query-conditioned planning and progressive tool instantiation rather than additional evidence, as analytic operations return compact results under the same budget.
Given its 11.3\% improvement in LLM-Judge accuracy over MM-RAG, the additional cost is acceptable for the resulting gains in analytical capability and answer quality.

\section{Case Study}
\label{app:case-study}

\begin{figure}[t]
\centering
\begin{tcolorbox}[
    enhanced,
    colback=gray!2,
    colframe=black!45,
    boxrule=0.45pt,
    arc=1mm,
    left=1.3mm,
    right=1.3mm,
    top=1mm,
    bottom=1mm
]
\footnotesize
\raggedright
\textbf{Query}\quad
On April 22, which screen, if any, shows both the highest Time In Range and the lowest Meal Variability?

\smallskip
\textbf{Planning context}\quad
\texttt{daily\_glucose\_dashboard} metadata and the \textsc{Rank} specification.

\smallskip
\tcbline
\smallskip
\textbf{High-level plan}
\begin{enumerate}[leftmargin=4.5mm,itemsep=0.15em,topsep=0.25em]
    \item $g_1$: Find the April 22 screen with the highest Time In Range. \hfill $\xi_1=\textsc{Rank}$
    \item $g_2$: Find the April 22 screen with the lowest Meal Variability. \hfill $\xi_2=\textsc{Rank}$
\end{enumerate}

\tcbline
\smallskip
\textbf{Progressive instantiation and execution}

\smallskip
\textbf{Step 1: highest Time In Range}\par
\begin{tabularx}{\linewidth}{@{}>{\raggedright\arraybackslash}p{0.14\linewidth}>{\raggedright\arraybackslash}X@{}}
$\Theta_1$ & \texttt{table}: \texttt{daily\_glucose\_dashboard};
\texttt{target\_column\_expression}: \texttt{time\_range};
\texttt{order}: descending;
\texttt{optional\_column\_filters}: \texttt{date = 2025-04-22};
\texttt{result\_budget}: 1. \\
$\Omega_1$ & \texttt{ranked\_rows}: \texttt{HEALTH\_S10:R5} (after-snack dashboard),
\texttt{time\_range = 90.0},
\texttt{meal\_variability = 28.0}. \\
\end{tabularx}

\smallskip
\textbf{Step 2: lowest Meal Variability}\par
\begin{tabularx}{\linewidth}{@{}>{\raggedright\arraybackslash}p{0.14\linewidth}>{\raggedright\arraybackslash}X@{}}
$\Theta_2$ & \texttt{table}: \texttt{daily\_glucose\_dashboard};
\texttt{target\_column\_expression}: \texttt{meal\_variability};
\texttt{order}: ascending;
\texttt{optional\_column\_filters}: \texttt{date = 2025-04-22};
\texttt{result\_budget}: 1. \\
$\Omega_2$ & \texttt{ranked\_rows}: \texttt{HEALTH\_S10:R7} (choice-comparison card),
\texttt{time\_range = 79.0},
\texttt{meal\_variability = 17.0}. \\
\end{tabularx}

\smallskip
\tcbline
\smallskip
\textbf{Answer}\quad
None. The highest Time In Range and lowest Meal Variability occur on different screens. \textcolor{green}{{(Correct)}}
\end{tcolorbox}
\caption{\textbf{Query-processing case study.}
The planner generates two analytic goals and progressively instantiates their \textsc{Rank} calls under a shared date constraint.}
\label{fig:case-query-processing}
\end{figure}

\begin{figure*}[t]
\centering
\begin{subfigure}[t]{0.322\textwidth}
    \centering
    \includegraphics[width=\linewidth]{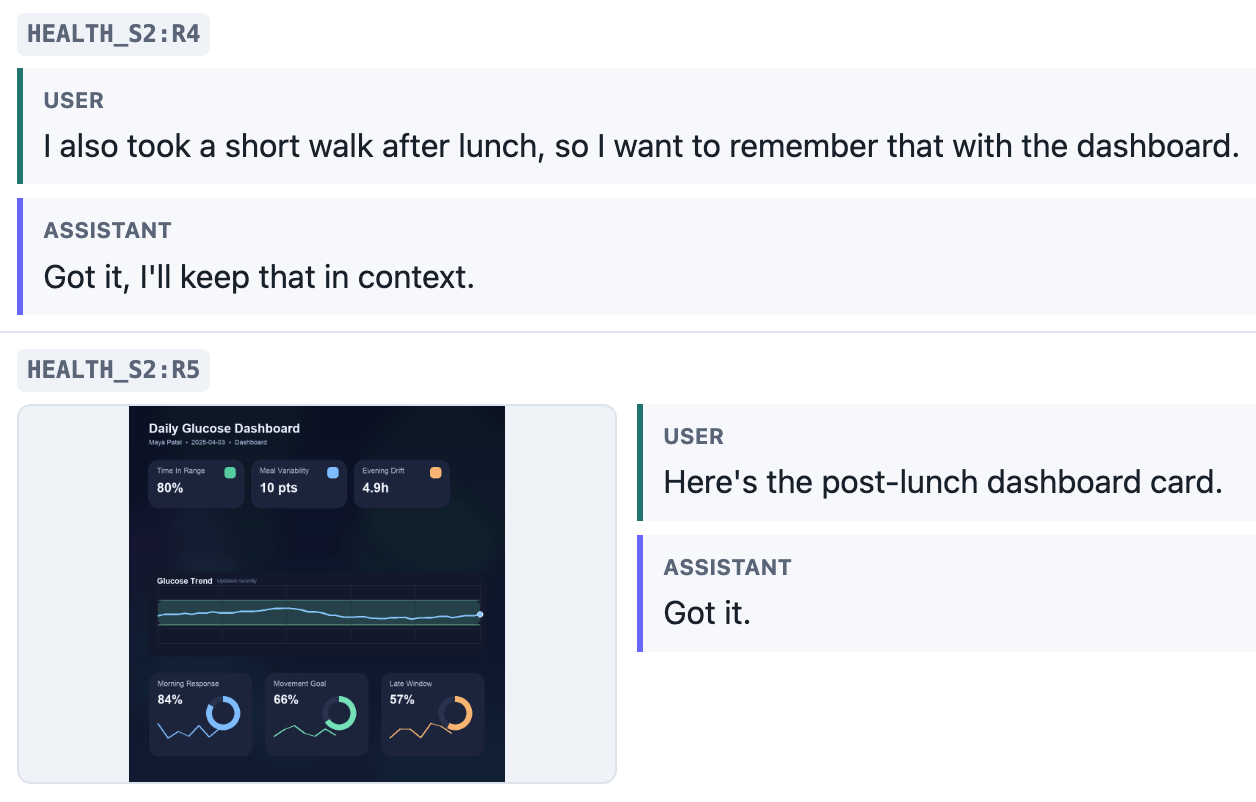}
    \caption{Session 2}
\end{subfigure}
\hfill
\begin{subfigure}[t]{0.322\textwidth}
    \centering
    \includegraphics[width=\linewidth]{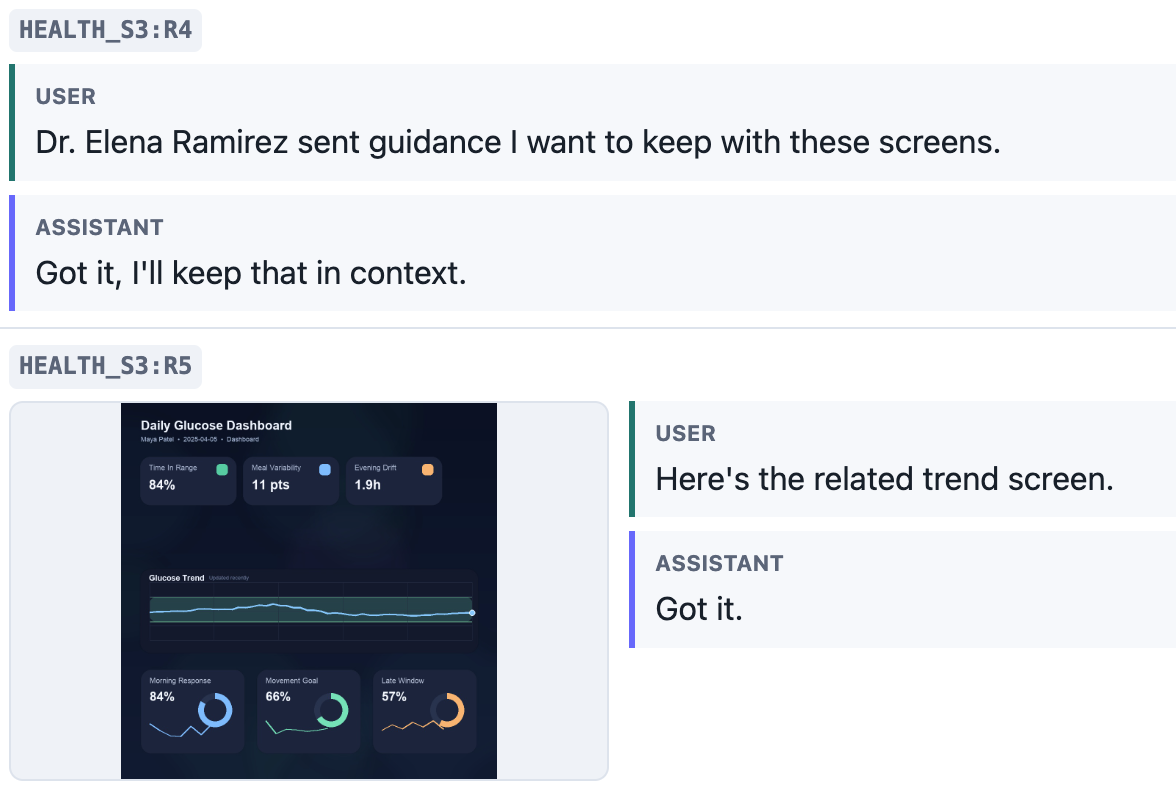}
    \caption{Session 3}
\end{subfigure}
\hfill
\begin{subfigure}[t]{0.322\textwidth}
    \centering
    \includegraphics[width=\linewidth]{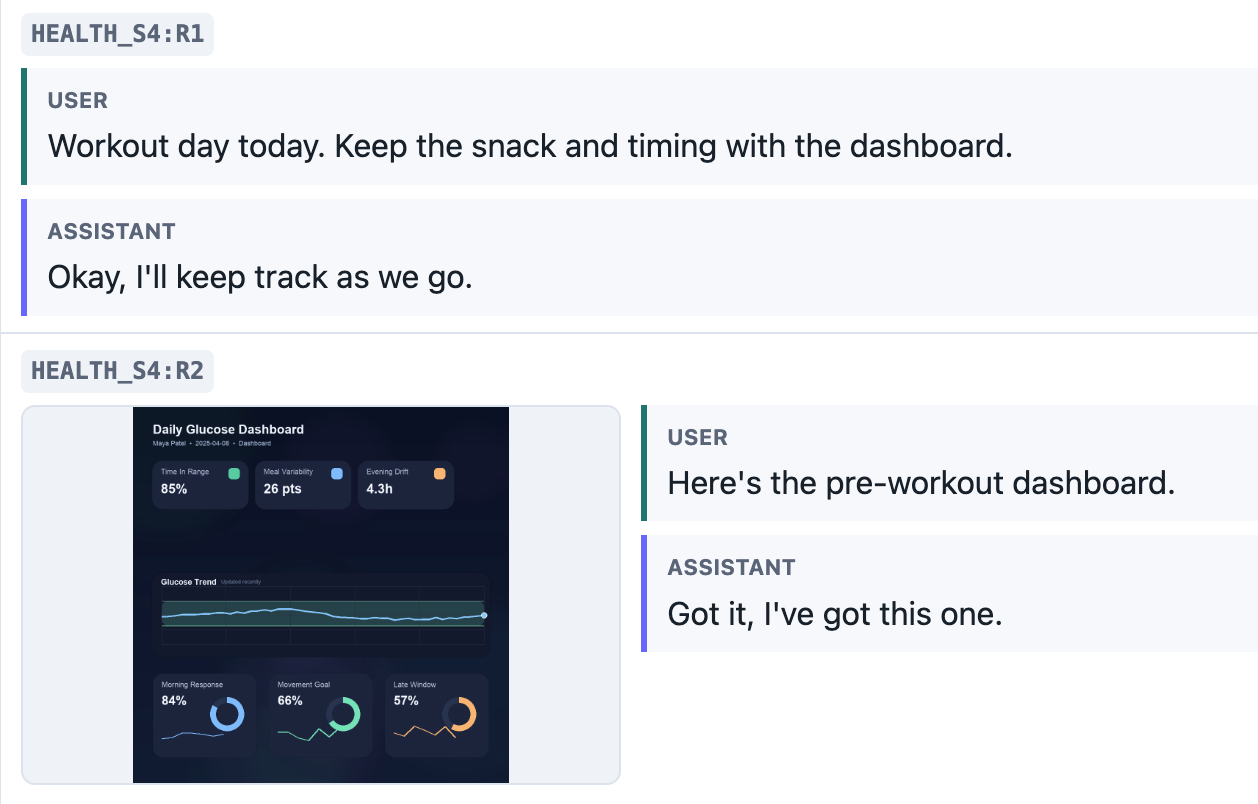}
    \caption{Session 4}
\end{subfigure}
\caption{Three representative multimodal interaction fragments sampled from the MemEye Health task.}
\label{fig:case-health-interactions}
\end{figure*}

\paragraph{Constructed analytic table.}
To illustrate the construction of analytic memory, Figure~\ref{fig:case-health-interactions} presents representative multimodal fragments from sampled Sessions 2, 3, and 4 of the MemEye Health task. 
For each interaction round, the LLM-based extractor first collects attribute--value observations from its dialogue and dashboard image, such as \texttt{time\_range}, \texttt{meal\_variability}, and \texttt{evening\_drift}. 
Candidate pattern mining then accumulates the occurrence and co-occurrence of these attributes across rounds. 
Although their values and surrounding activities vary, the dashboard attributes repeatedly occur together across rounds; once they satisfy the recurrence and co-occurrence-consistency criteria, schema induction identifies them as a shared analytic schema. 
During memory materialization, the schema attributes define stable columns, while the values extracted from each matched interaction populate one row. 
Finally, an LLM generates the table meta information, producing the \texttt{daily\_glucose\_dashboard} table, as shown in Table~\ref{tab:case-daily-glucose-dashboard}.

\begin{table*}[t]
\centering
\small
\caption{A representative analytic table constructed from the MemEye Health task. }
\label{tab:case-daily-glucose-dashboard}
\renewcommand{\arraystretch}{1.12}
\begin{tabularx}{\textwidth}{@{}>{\raggedright\arraybackslash}p{0.12\textwidth}>{\raggedright\arraybackslash}X@{}}
\toprule
\textbf{Table name}
& \texttt{daily\_glucose\_dashboard} \\
\textbf{Description}
& Longitudinal daily-glucose records that support exact comparison and aggregation of recurring glucose patterns, behavioral targets, and time-in-range trends across interactions. \\
\textbf{Columns}
& \texttt{source}, \texttt{round\_order}, \texttt{captions}, \texttt{date}, \texttt{evening\_drift}, \texttt{late\_window}, \texttt{meal\_variability}, \texttt{morning\_response}, \texttt{movement\_goal}, and \texttt{time\_range}. \\
\bottomrule
\end{tabularx}

\vspace{0.6em}
\scriptsize
\setlength{\tabcolsep}{2.2pt}
\begin{tabularx}{\textwidth}{
    @{}
    >{\raggedright\arraybackslash}p{0.102\textwidth}
    >{\centering\arraybackslash}p{0.050\textwidth}
    >{\raggedright\arraybackslash}X
    >{\centering\arraybackslash}p{0.060\textwidth}
    >{\centering\arraybackslash}p{0.052\textwidth}
    >{\centering\arraybackslash}p{0.070\textwidth}
    >{\centering\arraybackslash}p{0.070\textwidth}
    >{\centering\arraybackslash}p{0.070\textwidth}
    >{\centering\arraybackslash}p{0.055\textwidth}
    @{}
}
\toprule
\texttt{source} &
\texttt{round}\newline\texttt{\_order} &
\texttt{captions} &
\texttt{evening}\newline\texttt{\_drift} &
\texttt{late}\newline\texttt{\_window} &
\texttt{meal}\newline\texttt{\_vari...} &
\texttt{morning}\newline\texttt{\_response} &
\texttt{movement}\newline\texttt{\_goal} &
\texttt{time}\newline\texttt{\_range} \\
\midrule
\texttt{HEALTH\_S1:R2}
& 1
& Daily dashboard showing \ldots{} stable glucose trends
& 2.8 & 47.0 & 19.0 & 84.0 & 66.0 & 82.0 \\

\texttt{HEALTH\_S1:R3}
& 2
& A sleek daily \ldots{} time in range
& 4.3 & 57.0 & 16.0 & 84.0 & 66.0 & 80.0 \\

\texttt{HEALTH\_S1:R6}
& 5
& Daily Glucose Dashboard \ldots{} stable glucose trends
& 2.1 & 57.0 & 12.0 & 84.0 & 66.0 & 92.0 \\

\texttt{HEALTH\_S6:R5}
& 42
& Daily dashboard showing \ldots{} metrics for meals
& 2.2 & 57.0 & 27.0 & 84.0 & 66.0 & 90.0 \\

\multicolumn{9}{c}{\raisebox{0.1em}{$\boldsymbol{\cdots}$}} \\

\texttt{HEALTH\_S8:R6}
& 60
& A digital daily \ldots{} time in range
& 4.3 & 57.0 & 17.0 & 84.0 & 66.0 & 91.0 \\

\texttt{HEALTH\_S12:R8}
& 94
& A daily glucose \ldots{} time in range
& 1.3 & 57.0 & 19.0 & 84.0 & 66.0 & 82.0 \\
\bottomrule
\end{tabularx}
\end{table*}

\paragraph{Adaptive query processing.}
Figure~\ref{fig:case-query-processing} illustrates how the constructed table supports executable query processing.
Metadata retrieval first selects \texttt{daily\_glucose\_dashboard} and its applicable analytic tools.
Because the date is explicit, the high-level planner produces the shortest valid plan: two \textsc{Rank} operations for the maximum \texttt{time\_range} and minimum \texttt{meal\_variability} under the same filter.
The calls are progressively instantiated using the typed arguments in Table~\ref{tab:memory-access-tools}.
Since the two extrema occur in different rows, the answer model concludes that no screen satisfies both conditions.

\section{Prompt Templates}
\label{app:prompt-templates}

The key prompt templates used in \method are presented below.
Figures~\ref{fig:analytic-memory-prompts} and \ref{fig:planning-prompts} show the prompts for analytic memory construction and adaptive query processing, respectively.

\begin{figure*}[t]
\centering
\begin{AIbox}{Prompt templates for analytic memory construction.}
\textbf{Attribute extraction.}\\
You are a multimodal record extractor. Given one interaction round, identify every attribute--value observation explicitly supported by its dialogue, images, or contextual metadata. Do not assume a predefined attribute vocabulary, and do not infer facts absent from the input. Use concise attribute names, retain source-faithful values, and attach the dialogue-span or image identifier supporting each observation. Merge duplicate mentions within the round, but preserve distinct values.\\
Interaction round: \promptvar{Dialogue, contextual metadata, and images}\\
Interaction order: \promptvar{Order}\\
Timestamp: \promptvar{Time}\\
Output: \texttt{[\{"attribute": ..., "value": ..., "source\_type": ..., "source\_id": ...\}, ...]}

\medskip
\textbf{Table metadata generation.}\\
You are given the complete column set of a relational memory table and representative values sampled for each column. Generate a concise table name and a one-sentence description that distinguish the table from other memories. Preserve the provided column names without merging or renaming them, and infer each column type from \texttt{TEXT}, \texttt{INTEGER}, \texttt{REAL}, \texttt{BOOLEAN}, or \texttt{DATETIME}. The same procedure is used whenever a table is first materialized or its schema evolves.\\
Table columns: \promptvar{Complete column set}\\
Representative values: \promptvar{Sampled values by column}\\
Output: \texttt{\{"table\_name": ..., "description": ..., "columns": [\{"name": ..., "type": ...\}, ...]\}}
\end{AIbox}
\caption{Prompt templates for attribute extraction and table metadata generation.}
\label{fig:analytic-memory-prompts}
\end{figure*}



\begin{figure*}[t]
\centering
\begin{AIbox}{Prompt templates for adaptive query processing.}
\textbf{High-level plan generation.}\\
Given a user query and a planning context containing query-relevant memory metadata and access-tool specifications, produce the shortest executable plan that can obtain the information required to answer the query. Each step must specify one textual information goal and exactly one available tool. Use only the supplied memory candidates and tools, order dependent steps so that earlier results provide the information needed later, and use no more than three steps. Defer all concrete tool arguments to execution time, and do not answer the query.\\
Query: \promptvar{Query}\\
Planning context: \promptvar{Memory metadata and tool specifications}\\
Output: \texttt{[\{"goal": ..., "tool": ...\}, ...]}

\medskip
\textbf{Progressive tool instantiation.}\\
Instantiate the typed arguments for the designated tool at the current plan step. Use exact table, column, and memory identifiers from the planning context. When the step depends on an earlier result, derive its argument from the supplied preceding outputs rather than inventing a value. Respect the remaining evidence budget.\\
Query: \promptvar{Query}\\
Current information goal: \promptvar{Goal}\\
Designated tool: \promptvar{Tool name and argument schema}\\
Planning context: \promptvar{Memory metadata}\\
Preceding tool outputs: \promptvar{Previous results}\\
Remaining evidence budget: \promptvar{Budget}\\
Output: \promptvar{One JSON object matching the designated tool's argument schema}

\end{AIbox}
\caption{Prompt templates for high-level plan generation and progressive tool instantiation.}
\label{fig:planning-prompts}
\end{figure*}

\FloatBarrier

\end{document}